\documentclass[sigconf]{acmart}
\AtBeginDocument{%
  }

\copyrightyear{2026}
\acmYear{2026}
\setcopyright{cc}
\setcctype{by-nc-nd}
\acmConference[WWW '26]{Proceedings of the ACM Web Conference 2026}{April 13--17, 2026}{Dubai, United Arab Emirates}
\acmBooktitle{Proceedings of the ACM Web Conference 2026 (WWW '26), April 13--17, 2026, Dubai, United Arab Emirates}
\acmPrice{}
\acmDOI{10.1145/3774904.3793805}
\acmISBN{979-8-4007-2307-0/2026/04}

\usepackage{graphicx}
\usepackage{makecell}
\usepackage{siunitx}
\usepackage{calc}

\let\oldbibliography\thebibliography
\renewcommand{\thebibliography}[1]{%
  \oldbibliography{#1}%
  \setlength{\itemsep}{0pt}%
}

\begin{document}

\title[Linguistic Signatures of Emotion]{Linguistic Signatures for Enhanced Emotion Detection}


\author{Florian Lecourt}
\affiliation{%
  \institution{LIRMM, Université de Montpellier}
  \city{Montpellier}
  \country{France}
}

\author{Madalina Croitoru}
\affiliation{%
  \institution{LIRMM, Université de Montpellier}
  \city{Montpellier}
  \country{France}
}

\author{Konstantin Todorov}
\affiliation{%
  \institution{LIRMM, Université de Montpellier}
  \city{Montpellier}
  \country{France}
}

\renewcommand{\shortauthors}{Florian Lecourt, Madalina Croitoru, \& Konstantin Todorov}


\begin{abstract}
Emotion detection is a central problem in NLP, with recent progress driven by transformer-based models trained on established datasets. However, little is known about the linguistic regularities that characterize how emotions are expressed across different corpora and labels. This study examines whether linguistic features can serve as reliable interpretable signals for emotion recognition in text. We extract emotion-specific linguistic signatures from 13 English datasets and evaluate how incorporating these features into transformer models impacts performance. Our RoBERTa-based models enriched with high level linguistic features achieve consistent performance gains of up to +2.4 macro F1 on the GoEmotions benchmark, showing that explicit lexical cues can complement neural representations and improve robustness in predicting emotion categories.
\end{abstract}

\vspace{-0.3cm}


\vspace{-0.3cm}




\maketitle

\vspace{-0.3cm}

\section{Introduction}

Emotion recognition in text has become increasingly relevant for applications in social media analysis, dialog systems, and affective computing \cite{setyoningrum2025impact}. Transformer-based models such as BERT \cite{devlin2019bert} and RoBERTa \cite{liu2019roberta} have achieved strong results on established benchmarks \cite{adoma2020comparative}. In parallel, linguistic tools such as LIWC \cite{pennebaker2001liwc} and SEANCE \cite{crossley2017sentiment} capture interpretable lexical, cognitive, and affective features of language. However, explicit integration of hand-crafted or (psycho)linguistic features remains rare \cite{zanwar2022improving}. Moreover, probing analyses indicate that even when linguistic information is implicitly learned, many relevant features are not effectively utilized \cite{triantafyllopoulos2022probing}. This leaves open questions about how linguistic cues relate to model predictions \cite{kusal2023systematic}. This study addresses this gap by connecting linguistic and neural perspectives for emotion detection. This approach is particularly relevant for Web content analysis and mining. We are guided by two main research questions (RQ):  \looseness=-1 


\textbf{RQ1:} Can linguistic feature analysis across diverse benchmark datasets produce consistent and distinctive linguistic signatures for individual emotions?

\textbf{RQ2:} Do these linguistic signatures improve the performance of transformer-based emotion detection models?

To answer these questions, we extract linguistic signatures for each emotion label across thirteen publicly available English emotion datasets using SEANCE. We then integrate these features into transformer architectures leading to improved macro-F1 scores by up to +2.4 on the GoEmotions benchmark \cite{demszky2020goemotions}, and to effective generalizability, as evidenced by comparable performance to the State-of-the-art (SotA) on the CARER dataset \cite{saravia2018carer}. This indicates that linguistic cues are stable transferable signals for our task. \looseness=-1 

The paper is structured as follows: Section~\ref{sec:features} presents the dataset selection, linguistic feature extraction, and signature generation; Section~\ref{sec:models} introduces the RoBERTa- and LLaMA-based models incorporating linguistic features; Section~\ref{sec:discussion} discusses related work.

\vspace{-0.3cm}

\section{Dataset Selection and Linguistic Analysis} \label{sec:features}

To build a broad basis for linguistic analysis, we conducted a systematic survey of emotion-related datasets that have shaped research in Natural Language Processing (NLP) since the Transformers paper \cite{vaswani_2017_attention}. We used the keyword \textbf{emotion} to query leading NLP venues (ACL, EMNLP, COLING, etc.) from 2017 to 2024, retrieving 129 publications. After manual screening, 39 were deemed directly relevant to emotion detection, from which we extracted the datasets used for training or evaluation. Any dataset cited in these papers was also included to ensure comprehensive coverage. 

Because the vast majority of emotion datasets identified in our survey are in English, we restrict our study to English corpora. We further excluded resources that were inaccessible, non-standardized, or diverged from their documented format. As a result, 13 datasets remained, covering a wide range of textual contexts, from short social media posts to scripted dialogues, summarized in Table \ref{tab:datasetlist}.

\begin{table*}
\centering
\small 
\renewcommand{\arraystretch}{0.85} 
\begin{tabular}{| m{9em} | m{3em} | m{9em} | m{6em} | m{4em} | m{9em} | m{5em} | }
\hline
 \textbf{Title} & \textbf{Year} & \textbf{Source} & \textbf{Domain} & \textbf{Format} & \textbf{Size} & \textbf{\# of labels} \\
\hline
Affective Text \cite{strapparava2007semeval} & 2007 & News Headlines & News  & Snippets  & 1,250 headlines & 6 \\
\hline
BU-NEmo$^+$ \cite{gao2022prediction} & 2022 & News Headlines & News  & Snippets  & 1,297 headlines & 8 \\
\hline
Cancer Emo \cite{sosea2020canceremo} & 2020 & Online Cancer Survivors Network & Health & Snippets  & 25,000 online comments & 8 \\
\hline
CARER \cite{saravia2018carer} & 2018 & Twitter & Social Media & Snippets & 664,462 tweets & 6 \\
\hline
EmoContext \cite{chatterjee2019semeval} & 2019 & Conversation with Conversational Agent & Conversational Agent & Dialogues & 38,424 dialogues & 4 \\
\hline
EmoProgress \cite{wemmer2024emoprogress} & 2024 & Self-report & Psychological Self-report & Dialogues & 149 dreams + 339 dialogues & 10 \\
\hline
EmoryNLP \cite{zahiri2018emotion} & 2018 & Friends TV show & TV show & Dialogues & 897 scenes, forming 12,606 utterances & 7 \\
\hline
Empathetic Dialogues \cite{rashkin2018towards} & 2019 & ParlAI platform & Human-Human conversation & Dialogues & 24,850 conversations & 32 \\
\hline
GoEmotions \cite{demszky2020goemotions} & 2020 & Reddit & Social Media & Snippets & 58,009 Reddit comments & 28 \\
\hline
GoodNewsEveryone \cite{bostan2019goodnewseveryone} & 2019 & News Headlines & News & Snippets  & 10,470 headlines & 16 \\
\hline
IDEM \cite{prochnow_idem_2024} & 2024 & Generated with GPT-4 & Text Generation & Snippets & 9,685 sentences & 36 \\
\hline
ISEAR \cite{scherer_evidence_1994} & 1994 & Self-reporting statements & Psychological Self-report & Snippets & 7,666 texts & 7 \\
\hline
MELD \cite{poria2019meld} & 2019 & Friends TV show & TV show & Dialogues & 1,400 dialogues forming 13,000 utterances & 7 \\
\hline
\end{tabular}
\caption{Overview of the datasets}
\label{tab:datasetlist}
\end{table*}

SEANCE is based on the General Inquirer (GI) lexicon \cite{stone1966general}, which defines more than 180 semantically interpretable linguistic categories (e.g., Positiv, Legal, Virtue). The tool outputs normalized frequency values for each category and automatically accounts for negation. Comparative evaluations suggest that SEANCE performs on par with, and in some cases slightly outperforms, LIWC in capturing affective linguistic cues \cite{fiok2021analysis}. SEANCE has been used in prior work for large-scale discourse analysis in various domains, demonstrating its fitness for linguistic analysis \cite{bilotta2024examining}.

\begin{figure*}[h!]
    \centering
    \includegraphics[width=0.85\linewidth]{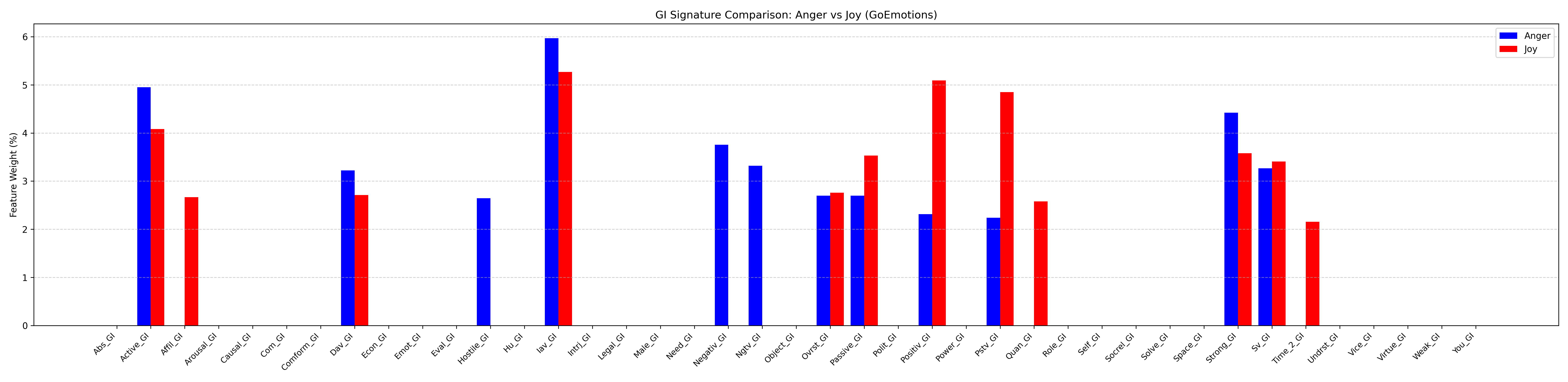}
    \vspace{-15pt}
    \caption{Comparisons of the signature for the Anger and Joy labels in the GoEmotions Dataset}
    \label{fig:signature_comparison}
\end{figure*}

To prepare data for feature extraction, all texts or utterances in each dataset were grouped by their corresponding emotion labels, with a dedicated plain-text file created for each emotion. In multi-label datasets, texts were duplicated under every relevant label to preserve co-occurrence information. SEANCE was then applied to each emotion-specific file, generating eigenvalues that capture the distribution of linguistic features across emotions. From these outputs, we retained the 10\% most frequent linguistic features per emotion, balancing noise reduction with coverage of salient emotional markers. These subsets define the \textbf{linguistic signatures} of emotions: compact representations of the linguistic categories most strongly associated with each label. (Cf. Figure \ref{fig:signature_comparison} for an example.)

To enable cross-dataset comparison, emotion labels were normalized by merging semantically equivalent categories (e.g., joy and joyful), yielding a harmonized set of 30 emotions. For each, we kept only GI features present in at least 50\% of instances to emphasize consistent rather than dataset-specific patterns.

We examined inter-emotion overlap using pairwise Jaccard similarity of their feature signatures. Among 435 emotion pairs, 60 showed strong overlap ($J > 0.7$), revealing recurring linguistic patterns. Features such as Active\_GI (words implying active orientation; e.g., swim, take), Iav\_GI (interpretative action verbs; e.g., encourage, flatter), and Strong\_GI (strength or power; e.g., achieve, great) appeared in over 90\% of emotions, whereas Virtue\_GI (moral approval or good fortune; e.g., honest, generous, unique to admiration), Hostile\_GI (hostility or aggressiveness; e.g., hate, attack, unique to anger), and Need\_GI (intent or desire; e.g., want, hedonistic, unique to desire) were emotion-specific. These linguistic signatures form a reproducible basis for cross-context emotion comparison. {No strong correlation emerged between linguistic features and dataset modality (snippets vs. dialogues) or domain (news, social media, TV transcripts).} The code and signatures are made available.\footnote{https://github.com/FlorianLecourt/Linguistic-Signatures-for-Enhanced-EmotionDetection} This analysis showed that consistent linguistic patterns can be identified across datasets (RQ1). The next section tests whether these patterns can inform transformer-based emotion models (RQ2).

\vspace{-0.3cm}

\section{Enhancing Emotion Detection Models} \label{sec:models}

We adopt RoBERTa due to its slight macro-F1 advantage over other transformers in multi-label emotion tasks \cite{cortiz2022exploring,paranjape2023converge}. All enhanced RoBERTa-based models were trained and evaluated on the GoEmotions corpus \cite{demszky2020goemotions}, to ensure comparability with the evaluation framework of \cite{kocon2023chatgpt}, who assessed ChatGPT across various NLP tasks, including emotion detection, reporting a BERT baseline with a macro F1 of 52.75 as the SOTA. GoEmotions was additionally chosen because it is a fine-grained, multi-label dataset covering 28 emotion categories, making it a more challenging and comprehensive benchmark compared to coarse-grained, single-label datasets. To verify reproducibility, we reimplemented their BERT model with the same preprocessing pipeline, obtaining 49.67 macro F1. This serves as our reproduction baseline. We normalized text following \cite{kodiyala2021emotion}: lowercase, hashtag and emoticon handling, and slang expansion to mitigate social-media noise. Training used five random seeds (1, 2, 10, 21, 42) with early stopping (patience = 3) and AdamW ($1{\times}10^{-5}$).

The first proposed model, RoBERTa-LexEnhance, is inspired by \cite{kodiyala2021emotion}. It augments RoBERTa by concatenating its [CLS] representation ($h_{\text{CLS}}$) with a SEANCE-derived vector ($s$) summarizing the relative frequency of each emotion’s General Inquirer (GI) categories in the input text. The final classifier input is the concatenation $[h_{\text{CLS}}; s]$, followed by a dropout layer ($0.2$) and a linear classifier.  

The second proposed model, RoBERTa-EarlyFusion, integrates lexical affective features at the token level, following  \cite{yang2024survey}. Each preprocessed sentence is tokenized, and every token is mapped to a binary vector encoding its associated GI categories. These token-level GI vectors are projected into an embedding space and transformed through a GELU-activated linear layer followed by a gating mechanism. The gate learns how strongly each GI feature should influence the base RoBERTa embedding. The resulting delta is scaled by a learned parameter $\alpha$ and added to the original embedding: $\tilde{E}_i = E_i + \alpha \, g_i \odot W_{\text{GI}}(x_i),$ where $E_i$ is the original embedding of token $i$, $x_i$ its GI feature vector, and $g_i$ the learned gate controlling attention to emotion feature. The modified embeddings $\tilde{E}$ are then processed by RoBERTa as usual, with a dropout of $0.3$. 

For completeness, we also evaluated a LLM baseline using LLaMA 3.2 \cite{touvron2023llama}. Each test instance from GoEmotions was processed via a structured prompting protocol requiring the model to output a Python list containing the exact number of ground-truth emotions. GI features and word lists per label were provided. 0-shot and 1-shot prompting conditions were tested. Invalid outputs were replaced with empty predictions. Results are given in Table~\ref{tab:model_results}.

\vspace{-0.2cm}

\begin{table}[H]
\centering
\small 
\renewcommand{\arraystretch}{0.85} 
\resizebox{\columnwidth}{!}{%
\begin{tabular}{|l|c|c|c|}
\hline
\textbf{Model} & \textbf{F1 (macro)} & \textbf{Precision} & \textbf{Recall} \\
\hline
SOTA baseline (BERT \cite{kocon2023chatgpt}) & 52.75 & N/A & N/A \\
Reproduction baseline & 49.67 & 51.87 & 48.57 \\
RoBERTa-LexEnhance (ours) & \textbf{55.18 ± 0.25} & \textbf{55.85 ± 0.50} & 57.60 ± 0.45 \\
RoBERTa-EarlyFusion (ours) & 54.98 ± 0.59 & 54.95 ± 0.77 & \textbf{57.79 ± 1.02} \\
LLaMA 3.2 (0-shot) & 0.23 & 9.52 & 0.12 \\
LLaMA 3.2 (1-shot) & 5.01 & 21.51 & 3.31 \\
\hline
\end{tabular}%
}
\caption{Overall results on the GoEmotions dataset.}
\label{tab:model_results}
\end{table}

\vspace{-1.0cm}

\begin{table}[H]
\centering
\small 
\renewcommand{\arraystretch}{0.85} 
\resizebox{\columnwidth}{!}{%
\begin{tabular}{|l|c|c|c|}
\hline
\textbf{Model} & \textbf{F1 (macro)} & \textbf{Precision} & \textbf{Recall} \\
\hline
SOTA baseline (BERT \cite{prasadm2022Emotiondetectionfromtext}) & \textbf{90.00} & \textbf{90.00} & 90.00 \\
RoBERTa-LexEnhance (ours) & 88.68 & 88.15 & 90.08 \\
RoBERTa-EarlyFusion (ours) & 88.52 & 87.35 & \textbf{90.18} \\
\hline
\end{tabular}%
}
\caption{Overall results on the CARER dataset.}
\label{tab:model_results_Carer}
\end{table}

\vspace{-0.8cm}

Both RoBERTa-based variants surpassed the SOTA baseline. RoBERTa-LexEnhance achieved the best macro-F1 and precision, while RoBERTa-EarlyFusion reached the highest recall.   
LLaMA 3.2 underperforms drastically, likely due to prompt overload: its complex instructions may have hindered consistent mapping between features, words, and emotion labels.

To complete our analysis, both RoBERTa-based models were fine-tuned on the CARER dataset \cite{saravia2018carer}, a coarse-grained single label dataset, and compared to its SOTA baseline (Table \ref{tab:model_results_Carer}). Performance was near-identical to SOTA overall, indicating strong transfer from GoEmotions. EarlyFusion exceeded SOTA in global recall and on Love, while LexEnhance surpassed it on Anger. Both models outperformed SOTA on Sadness but lagged on Surprise.

\begin{figure*}[t]
    \centering
    \includegraphics[width=0.85\linewidth]{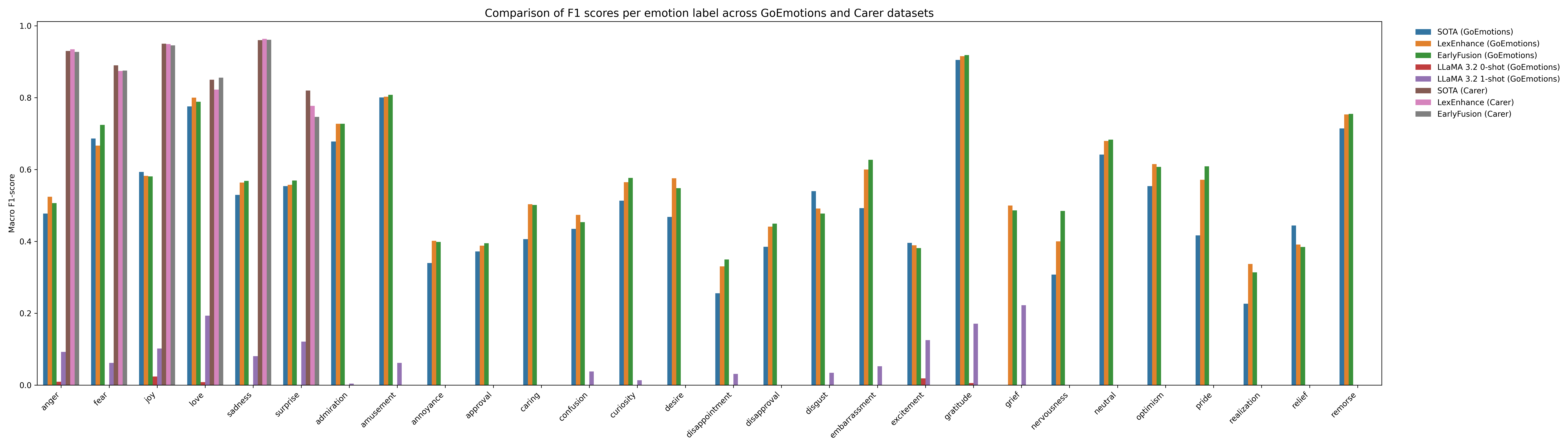}
    \vspace{-15pt}
    \caption{Comparison of F1 scores per emotion label across GoEmotions and Carer datasets.}
    \label{fig:class_f1_comparison}
\end{figure*}

Figure \ref{fig:class_f1_comparison} compares F1-scores per emotion across models, evaluating performance on the GoEmotions benchmark and transferability to the CARER dataset. On GoEmotions, except for Disgust, Excitement, Joy, and Relief, at least one of our models surpasses the SOTA. Grief, undetected by SOTA, was reliably identified by both our RoBERTa variants. LexEnhance excels for semantically well-defined emotions (e.g., admiration, anger, desire), which are strongly anchored in lexical semantics and tend to have clear linguistic markers \cite{bann2012discoveringbasicemotionsets}.
In contrast, EarlyFusion performs better for context-sensitive emotions (e.g., embarrassment, pride), whose interpretation depends on situational or social cues \cite{flink2019responding, bonanno2013regulatory}. This complementarity shows that global feature aggregation enhances precision, while token-level fusion increases sensitivity. Overall, we show that incorporating linguistic features enhances cross-emotion generalization and robustness for underrepresented emotion classes across datasets, answering RQ2. 

\vspace{-0.2cm}

\section{Related Work} \label{sec:discussion}

Previous research has sought to bridge symbolic and neural approaches to emotion and sentiment modeling.  
Mestre-Mestre analyzed affective framing in political communication using SEANCE features, showing how lexical indices can capture consistent variations in emotional tone across time and topics \cite{mestre2023emotion}.  
Similarly, Mohammad and Turney’s NRC Emotion Lexicon mapped words to Plutchik’s emotion wheel \cite{plutchik_emotions_1991}, enabling feature-level supervision for sentiment models \cite{mohammad2013nrc}.  
However, such lexicon-based studies have been primarily descriptive and applied to sentiment analysis or domain-specific contexts such as health, politics, or social media communication and rarely address fine-grained emotion detection.

Efforts to integrate affective knowledge into neural models have followed two main strategies.  
Ke et al. introduced SentiLARE, which injects linguistic knowledge from SentiWordNet during pre-training via a label-aware masked language modeling objective, effectively encoding sentiment polarity and part-of-speech information at the embedding level \cite{ke2020sentilare}.  
Kodiyala and Mercer proposed an alternative fine-tuning strategy that augments BERT with polarity-based lexicon features concatenated to sentence embeddings, improving classification robustness on Twitter datasets \cite{kodiyala2021emotion}.  
Despite these advances, such approaches rely on sentiment-oriented resources.  

Our work differs by grounding model enhancement in the GI framework, thereby introducing cognitively interpretable features into transformer pipelines.  
This design not only improves predictive performance but also allows direct comparison between computational representations and human affective theory.

\vspace{-0.3cm}

\section{Conclusion}
Leveraging a linguistic analysis of a large pool of emotion detection datasets, we showed how linguistic features can be used to characterize emotion labels and hence enhance emotion detection. Future work will focus on developing a quantitative emotion-signature metric that measures the proximity of emotions based on their linguistic profiles, enabling systematic comparison across corpora. While we focus here on English to ensure consistent feature extraction via SEANCE, we acknowledge that this restricts generalizability; subsequent studies should extend this methodology to multilingual Web content and explore Chain-of-Thought prompting to better leverage linguistic cues.

\vspace{-0.2cm}

\begin{acks}
This research was supported by the European Regional Development Fund (FEDER) through the IA-EMOTION project.
\end{acks}

\vspace{-0.3cm}

\bibliographystyle{ACM-Reference-Format}
\bibliography{sample-base}

\end{document}